%% file: main.tex
\pdfoutput=1
\documentclass[11pt]{article}

\usepackage[final]{acl}

\usepackage{times}
\usepackage{latexsym}
\usepackage{multirow}
\usepackage{amsmath}
\usepackage{booktabs}
\usepackage{xcolor}
\usepackage{tabularx}
\usepackage{graphicx}
\usepackage{changes}
\usepackage[T1]{fontenc}
\usepackage[utf8]{inputenc}

\usepackage{microtype}

\usepackage{inconsolata}

\usepackage{algorithm}
\usepackage{algorithmic}

\usepackage{tcolorbox}
\usepackage{array}
\usepackage{lipsum} % 用于生成示例文本，可以根据需要去掉
\usepackage{arydshln}
\usepackage{float}
\definecolor{lightgray}{gray}{0.9}

\title{Retrieve-Plan-Generation: An Iterative Planning and Answering Framework for Knowledge-Intensive LLM Generation}

\newcommand*{\affaddr}[1]{#1}
\newcommand*{\affmark}[1][*]{\textsuperscript{#1}}
\newcommand*{\email}[1]{\texttt{#1}}

\author{
Yuanjie Lyu\affmark[\textnormal{1}]$^{\dagger}$, Zihan Niu\affmark[\textnormal{1}]$^{\dagger}$, Zheyong Xie\affmark[\textnormal{1}], Chao Zhang\affmark[\textnormal{1}],
\\ \bf Tong Xu\affmark[\textnormal{1}]\thanks{Corresponding author.}, Yang Wang\affmark[\textnormal{2}], Enhong Chen\affmark[\textnormal{1}]\\
\affaddr{\affmark[1]University of Science and Technology of China} \\
\affaddr{\affmark[2]Anhui Conch Information Technology Engineering Co., Ltd.} \\
\email{\{s1583050085, zclfe00\}@gmail.com,}\\
\email{\{niuzihan, xiezheyong\}@mail.ustc.edu.cn, } \\
\email{wangyang@chinaconch.com,}\\
\email{\{tongxu, cheneh\}@ustc.edu.cn} \\
}
\begin{document}
\maketitle
\renewcommand{\thefootnote}{\fnsymbol{footnote}}
\footnotetext[2]{Equal Contribution. }
\renewcommand{\thefootnote}{\arabic{footnote}}
\begin{abstract}
Despite the significant progress of large language models (LLMs) in various tasks, they often produce factual errors due to their limited internal knowledge. Retrieval-Augmented Generation (RAG), which enhances LLMs with external knowledge sources, offers a promising solution. However, these methods can be misled by irrelevant paragraphs in retrieved documents. Due to the inherent uncertainty in LLM generation, inputting the entire document may introduce off-topic information, causing the model to deviate from the central topic and affecting the relevance of the generated content.
To address these issues, we propose the Retrieve-Plan-Generation (RPG) framework. RPG generates plan tokens to guide subsequent generation in the plan stage. In the answer stage, the model selects relevant fine-grained paragraphs based on the plan and uses them for further answer generation. This plan-answer process is repeated iteratively until completion, enhancing generation relevance by focusing on specific topics. To implement this framework efficiently, we utilize a simple but effective multi-task prompt-tuning method, enabling the existing LLMs to handle both planning and answering. We comprehensively compare RPG with baselines across 5 knowledge-intensive generation tasks, demonstrating the effectiveness of our approach.
\footnote{Our code and models will be publicly available at \href{https://github.com/haruhi-sudo/RPG}{https://github.com/haruhi-sudo/RPG}.}
\end{abstract}

\section{Introduction}
\input{content/introduction}

\section{Related Work}

\input{content/relatedwork}

\section{Methodology}
\input{content/method}

\section{Experiments}

\input{content/experiments}

\section{Conclusion}
\input{content/conclusion}

\section{Limitations}
\input{content/limitations}

% \balance
\bibliography{anthology}

\clearpage
\appendix
\input{content/appendix}

\end{document}

%% file: content/introduction.tex
\begin{figure}[t]
\centerline{\includegraphics[width=0.49\textwidth]{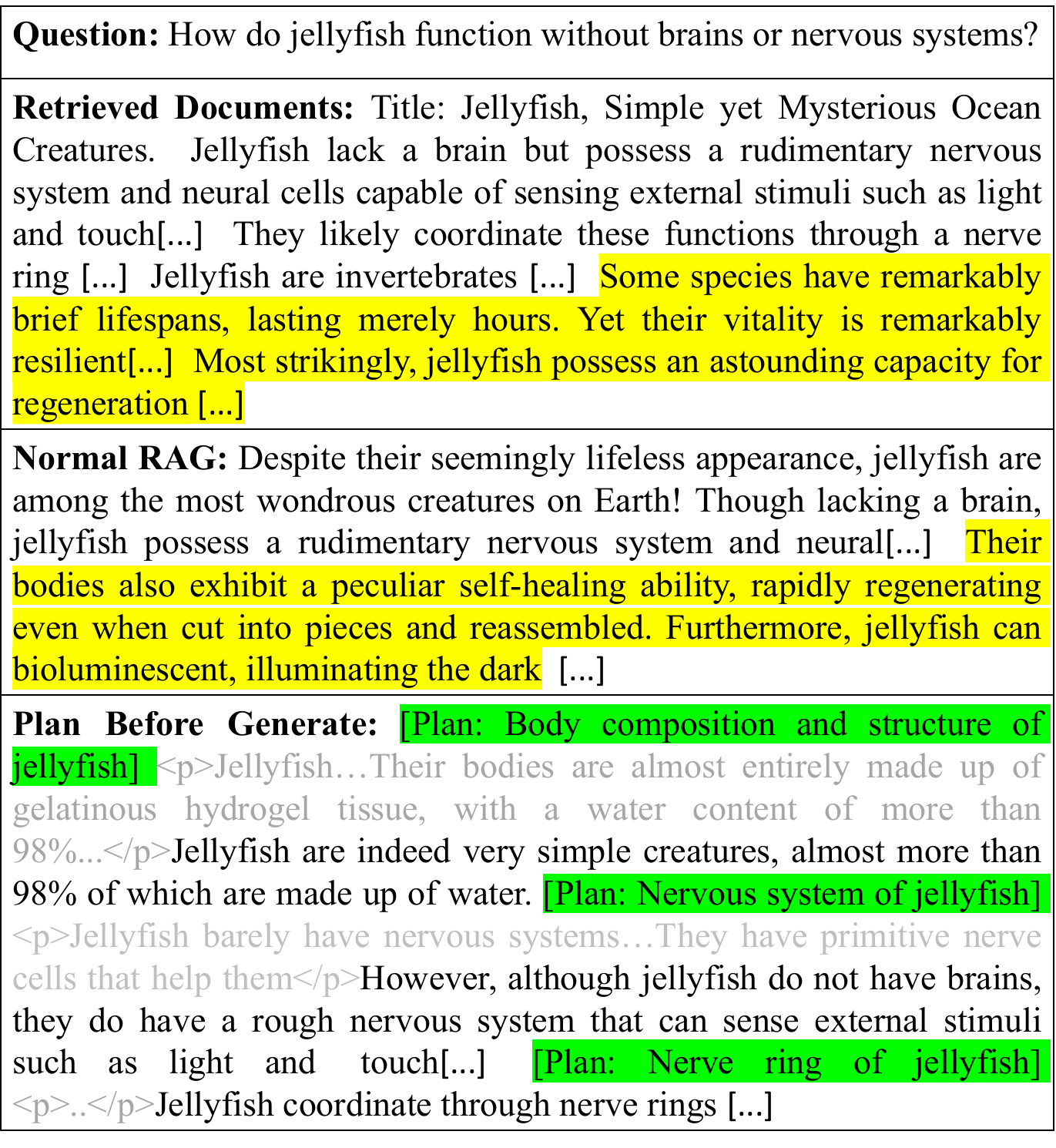}}
\caption{The retrieval documents contain off-topic paragraphs (highlighted in yellow), causing potential deviations in RAG outputs. By planning first (highlighted in green), selecting relevant fine-grained paragraphs, and then answering, the plan-answer iteration ensures a more consistent and relevant generation.
}
\label{figure:intro}
\end{figure}

With the persistent scaling up of training parameters and datasets~\cite{kaplan2020scaling}, large language models (LLMs)~\cite{touvron2023llama,jiang2023mistral,bai2023qwen,achiam2023gpt} have made remarkable advancements, becoming the cornerstone of many Natural Language Processing (NLP) tasks in recent years. Despite improvements in model architecture and the expansion of training data, 
LLMs still struggle with factual errors~\cite{lyu2022faithful,he2022rethinking}. To address this issue, the Retrieval-Augmented Generation (RAG) system has been introduced~\cite{lewis2020retrieval, guu2020retrieval}. By retrieving external information and incorporating it into the input, the RAG system demonstrates excellent performance in knowledge-intensive tasks.

The most common approach in RAG involves using the user input as a query for a single-time retrieval~\cite{lewis2020retrieval}, with LLMs then generating answers based on the retrieved information. However, documents retrieved for input into the LLM are often lengthy, and not all paragraphs may be practically helpful for answering the question.  Recent research~\cite{lan2021modeling,sun2023towards} indicates that off-topic paragraphs can be detrimental to the generation. As Figure ~\ref{figure:intro} illustrates, due to the inherent uncertainty in the generation process of LLMs, inputting the entire retrieved document can lead to those off-topic paragraphs misleading the model, causing a shift in focus and resulting in content that gradually deviates from the main topic.

Currently, many researchers have acknowledged this issue and have adopted various solutions. Some works~\cite{jiang2023active, asai2023self} determine whether retrieval is necessary before generating an answer and input the retrieved document only when required. Self-RAG~\cite{asai2023self} further introduces reflection tokens to evaluate the quality of retrieved documents, thereby excluding irrelevant documents. Despite significant advancements with these methods, their effectiveness diminishes when dealing with longer retrieved texts, particularly those that are generally relevant but contain some irrelevant details. Additionally, when the retrieved documents are too lengthy, it becomes challenging for users to verify the correctness of specific details in the generated content.

We propose that the susceptibility of LLMs to irrelevant content stems from a lack of explicit pre-planning in generating subsequent content. As illustrated in Figure ~\ref{figure:intro}, if the model continuously plans the next topic at each step and only focuses on highly relevant paragraphs, it can avoid being misled by irrelevant material during lengthy generation processes. To implement this plan-answer process, we introduce the Retrieve-Plan-Generation (RPG) framework. RPG iterates through two stages: the \textbf{\textit{plan stage}} and the \textbf{\textit{answer stage}}. In the plan stage, the model generates tokens representing upcoming text topics. During the answer stage, the model selects highly on-topic paragraphs from retrieved documents based on these topics, and uses them to generate targeted answers. This iterative process between planning and answering continues until the generation is complete. Unlike traditional full-text input methods, RPG provides detailed control over content generation by focusing on specific topics at each step, ensuring the generation is highly relevant and accurate. Additionally, this fine-grained approach makes it easier for users to verify the correctness of answer details, even when dealing with long documents.

Existing LLMs struggle to effectively integrate both planning and answering capabilities. 
Since the plan must be incrementally developed during the generation process, relying solely on pre-designed prompts for plan generation is challenging. Additionally, prompts need to guide the model in generating both the plan and the answer based on generated context and relevant paragraphs, which imposes high demands on the model's ability to comprehend complex prompts.
Therefore,
we prompt ChatGPT to create supervision for plan generation and fine-grained paragraph utilization based on existing datasets~\cite{asai2023self, yang2018hotpotqa}, then train our model end-to-end on this dataset. 

Fully fine-tuning an LLM is resource-intensive and often unnecessary.
% ~\cite{lester2021power, ding2022delta}
To balance the learning capabilities of the LLMs with training efficiency, prompt tuning has emerged as a promising method. Given that the input and output formats for planning and answering tasks differ, we adopt a multi-task prompt tuning approach, training two learnable prompt tokens specifically for plan and answer generation. 
These two prompt tokens share the same soft prompt. 
During the training stage, each data is simultaneously used for both planning and answering tasks. To train task-specific prompts, we first transform the soft prompt to the corresponding task mode, and then exclude the impact of other parts during loss computation.

Empirical results on 5 tasks, including long-form, multi-hop, and short-form generation, demonstrate that RPG significantly outperforms instruction-tuned LLMs with more parameters and widely adopted RAG approaches. Technical contributions of this paper can be summarized as follows:

\begin{figure*}[htbp]
\centerline{\includegraphics[width=1\textwidth]{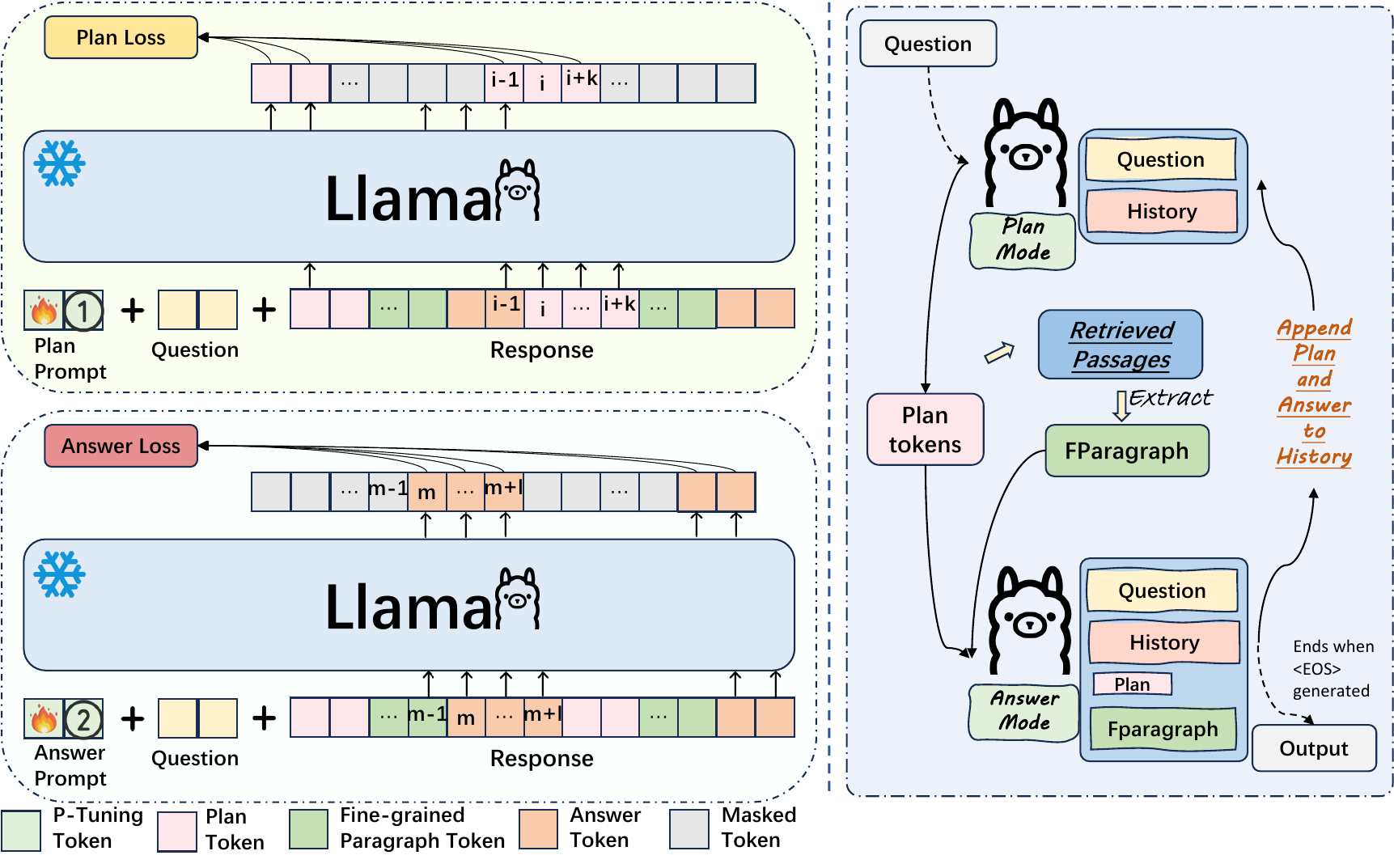}}
\caption{
Illustration of the proposed RPG. The left shows the training process, where plan and answer tasks use the same example data, different loss functions, and train two task-specific prompts simultaneously. The right shows the inference process, where the plan-answer process is repeated iteratively until completion.
}
\label{figure:model}
\end{figure*}

\begin{enumerate}
\item[\textbullet]We propose a new framework, RPG, which incorporates an explicit planning stage for LLMs, enhancing generation relevance by focusing on specific topics iteratively.

\item[\textbullet]We also adopt a simple but effective method that enables existing LLMs to easily configure plan-answer capabilities, adapting to the distinct requirements of these two tasks through multi-task learning.

\item[\textbullet] Experimental results on 5 tasks demonstrate the superiority of our proposed method over state-of-the-art methods.
\end{enumerate}

%% file: content/relatedwork.tex
\textbf{Retrieval-Augmented Generation.} Retrieval-Augmented Generation (RAG)~\cite{lewis2020retrieval, guu2020retrieval} enhances LLMs by retrieving relevant passages, thereby improving both the quality and accuracy of generated content, particularly in knowledge-intensive tasks~\cite{shen2023chatgpt,chen2023hallucination}. Early works~\cite{es2023ragas,lyu2024crud} chose to retrieve once, incorporating a fixed number of retrieved passages with a query into LLMs to generate a response. Recent research indicates that adaptive retrieval, tailored to the demands of LLMs, can further enhance generation. FLARE~\cite{jiang2023active} uses the generated sentence with a low confidence score as the query to retrieve external knowledge adaptively and then regenerates the current sentence, while Self-RAG~\cite{asai2023self} introduces special tokens allowing the model to adaptively retrieve and reflect the quality of generated content. SuRe~\cite{kim2024sure} generates conditional summarizations of retrieval and evaluating them with carefully designed prompts. However, existing approaches may not take full advantage of the planning capabilities of LLMs. Additionally, these methods may struggle to extract relevant content from retrieved passages and are easily influenced by irrelevant information. 

 \noindent\textbf{Parameter-Efficient Fine-Tuning.}  
Despite the powerful generative capabilities of LLMs, fine-tuning them requires substantial computational resources~\cite{lester2021power, ding2022delta, liu2023gpt}. To achieve more efficient fine-tuning, parameter-efficient tuning methods have emerged. These methods either fine-tune a small portion of the model parameters or introduce additional learnable parameters without fine-tuning the model itself~\cite{hu2021lora, liu2021p, ding2022delta, wang2023multitask}. LoRA (Low-Rank Adaptation)~\cite{hu2021lora} reduces the number of parameters to be updated by decomposing the weight matrices into low-rank components. Prompt tuning~\cite{liu2021p, liu2023gpt} introduces task-specific prompts by concatenating learnable tokens before the input sequence, requiring minimal parameter updates. Multi-task Prompt Tuning (MPT)~\cite{wang2023multitask} further highlights the commonalities between multi-task learning, suggesting that using a shared soft prompt and task-specific low-rank matrices can yield better results. 

%% file: content/method.tex
In this section, we first introduce the task definition and basic notation. Then, we provide a comprehensive explanation of the RPG framework from the perspectives of fine-grained dataset construction, training, and inference.

\subsection{Task Definition \& Notation} 
Given a user input $x$, a retriever $\mathcal{R}$ and document corpus $\mathcal{D}=\{d_1, d_2, \dots, d_n\}$, RAG aims to enhance the quality of a language model's (LM's) output $y$ by retrieving relevant passages from $\mathcal{D}$ and incorporating them into the answer. For a query $q$, the retriever $\mathcal{R}$ can retrieve a list of documents $\mathcal{D}_q = \mathcal{R}(q, \mathcal{D})$ from corpus $\mathcal{D}$.

\noindent\textbf{Vanilla Retrieval Augmented Generation.} The most common approach is to use the user input $x$ directly as the query for retrieval, and then generate the complete answer in a single step $y = LM([\mathcal{D}_x, x])$. 

\noindent\textbf{Dynamic Retrieval Generation.} 
To aid long-form generation with retrieval, 
dynamic retrieval generation further refines the RAG approach by dynamically retrieving information according to the model's needs during the generation process. 
Although dynamic retrieval can reduce the factual errors of LM, the lack of explicit planning may lead to interference from irrelevant information, resulting in the focus shift phenomenon.
Based on this fundamental structure, this paper innovatively proposes a two-stage method using the distinct plan and answer stage to achieve generated content with reduced focus shift.

\subsection{Method Overview}
To enhance the factuality of LLMs and improve topic consistency in long-form generation, LLMs should be capable of generating a preliminary plan to select fine-grained evidence, guiding subsequent content generation on specific topics.
Based on this consideration, our RPG framework is designed into two stages: \textit{\textbf{plan}} and \textit{\textbf{answer}}. 
During the plan stage, the LLM should generate a topic for the upcoming answer, reflecting pre-planned thoughts and guiding the subsequent generation. This approach effectively prevents the output from deviating from the specific topic. In the answer stage, by removing irrelevant information at the sentence level, a foundational denoising capability is achieved. This decouples the processes of filtering and utilizing relevant information during the generation, thereby enhancing the model's ability to leverage fine-grained relevant evidence. Through the iterative alternation of these two stages, the focus shift phenomenon during long text generation can be effectively avoided.
Specifically, to train an LLM end-to-end with both planning and fine-grained evidence utilization capabilities efficiently, multi-task prompt tuning is employed to learn these two tasks synchronously on a dataset we reconstructed.
During the inference stage, the LLM iteratively repeats the plan-answer process until the final response is generated.
Figure~\ref{figure:model} illustrates both the training and inference of the RPG framework.

\subsection{Dataset Construction}
To train the aforementioned LLM, we reconstruct a fine-grained dataset based on the existing Self-RAG~\cite{asai2023self} and HotpotQA~\cite{yang2018hotpotqa} datasets, where the annotated data has been split into segments with retrieved documents. 

\begin{figure}[t]
\centering
\includegraphics[width=\columnwidth]{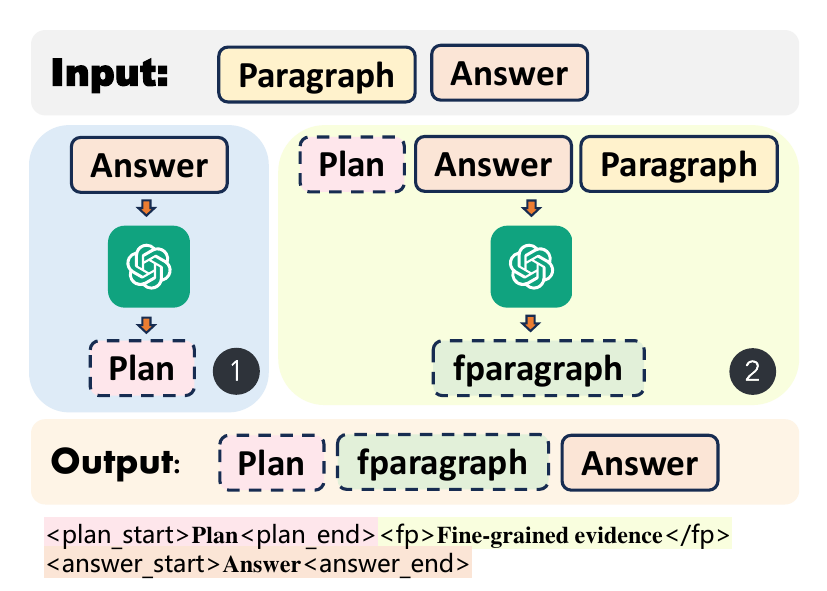}
\caption{Illustration of the data processing for one of the segments in a sample. 
% The basic composition of processed data is shown below the figure.
}
\label{figure:data}
\end{figure}

\noindent\textbf{Data collection for plan.} 
Since answer segments are specific implementations of an individual's planning at each step, we can treat the intent of these segments as human planning, thereby avoiding topic deviation. As shown in Figure \ref{figure:data}, to obtain the intent of each segment, we prompt ChatGPT to summarize the segment and use these summaries as labels for the plan stage. For data that do not require additional retrieved information, we attach \texttt{<\textsc{not\_need\_extra\_info}>} directly at the start of the answer, indicating no planning is needed and the LLM's inherent ability to answer is sufficient.
 
\noindent\textbf{Data collection for answer.} 
As mentioned before, the coarse-grained documents provided in existing datasets often contain off-topic paragraphs, which has been shown to be adverse to generation~\cite{yoran2023making}. 
After filtering the paragraphs at the sentence level, we retain only the information related to the plan tokens and the corresponding answer segment for the answer stage training.
Specifically, we provide ChatGPT with pre-generated plan tokens, along with corresponding coarse-grained documents and the answer segment. We then require ChatGPT to select sentences related to the plan and answer from the document as fine-grained evidence, which is further used to train the LLM's ability of fine-grained evidence utilization. The answer segments are the labels for the answer stage.

Finally, we collect 50k supervised training data to form a new dataset for RPG training. More details about our dataset are shown in Appendix~\ref{appendix:dataset statistical}. Prompts and examples are shown in Appendix ~\ref{appendix:prompts}.
% \vspace{2mm}
\subsection{RPG Training}

To efficiently leverage the information within the data, we introduce a multi-task training method for the RPG framework. During the training phase, we utilize different components of the samples, plan and answer, from the constructed dataset to train the model. Simultaneously, we train two task-specific learnable prompts with different loss functions. This approach enables the frozen LLM to acquire planning and answering capabilities without requiring any modifications to the model itself.

As shown in Figure~\ref{figure:model}, 
to achieve more parameter-efficient fine-tuning,  we opt to freeze the LLM and train the additional continuous prompt vectors prepended to the input. Recent research~\cite{wang2023multitask} indicates that commonalities exist across various tasks, paving the way for more efficient prompt tuning. Following them, we first employ a soft prompt $P^{*}$ as the shared prompt across plan and answer tasks. To adapt to the distinct requirements of these two tasks, we further utilize two different low-rank matrices, $W_{plan}$ and $W_{ans}$, to transform the soft prompt to the specific task mode. The task prompts for plan and answer generation task are parameterized as follows:
\begin{equation}
    \begin{aligned}
        P_{task} = P^{*} \circ W_{task} = P^{*} \circ (u_{task} \otimes v_{task}^T),
    \end{aligned}
\end{equation}
where $\circ$ denotes the Hadamard product between two matrices, and $task \in \{plan, ans\}$ denotes the specific generation task.

To enhance the efficiency of multi-task training, we utilize different components of the samples to simultaneously train the plan prompt and the answer prompt. Specifically,  we adopt to mask different parts of the same data instance to guide the learning of corresponding tasks. For the \textbf{\textit{plan stage}} training, tokens other than the plan tokens in the ground truth are masked, guiding the LLM to focus solely on plan generation. Similarly, for the \textbf{\textit{answer stage}}, tokens that are not part of the answer are excluded from the loss calculation.  For formal expression, the conditional language modeling objective $\mathcal{L}_{plan}$ and $\mathcal{L}_{ans}$ are employed to optimize our model $\mathcal{M}$ in two stages:
\begin{equation}
    \begin{aligned}
        \mathcal{L}_{plan} = -\sum_{y_i \in plan} \log P(y_i|x_i; \Theta, P_{plan}), 
    \end{aligned}
\end{equation}
\begin{equation}
    \begin{aligned}
        \mathcal{L}_{ans} = -\sum_{y_i \in ans} \log P(y_i|x_i; \Theta, P_{ans}),
    \end{aligned}
\end{equation}
where $P_{plan}$ and $P_{ans}$ are learnable. During training, we combine the two loss functions and optimize the model parameters simultaneously.

\subsection{RPG Inference}

\begin{algorithm}
    \caption{RPG Inference}
    \begin{algorithmic}
        \REQUIRE{Generator LLM $\mathcal{M}$, Retriever $\mathcal{R}$, Large-scale passage collections $\mathcal{D}=\{d_1,\dots,d_N\}$, Task Prompts $P_{{plan}}$, $P_{{ans}}$}
    \end{algorithmic}
    \begin{algorithmic}[1]
         \STATE \textbf{Input:} user input $x$ and retrieved relevant passages $\mathcal{D}_x = \mathcal{R}(x, \mathcal{D})$, \textbf{Output:} response $y$
         \STATE Initialize the response $y \leftarrow \emptyset$
         \STATE $\mathcal{M}$ predicts plan $\mathcal{P}$ given $(P_{{plan}}, x)$
         \IF{$\mathcal{P} == \texttt{<\textsc{not\_need\_extra\_info}>}$}
         \STATE $\mathcal{M}$ generates $y$ given $(P_{ans}, x)$
         \ELSE
         \WHILE{$\mathcal{M}$ has not generated the \texttt{<\textsc{EOS}>} token}
         \STATE Select relevant paragraphs $e$ given $(\mathcal{D}_x, \mathcal{P})$
         \STATE $\mathcal{M}$ predicts $y_t$ given $(P_{ans}, x, e, y_{<t}, \mathcal{P})$
         \STATE $\mathcal{M}$ predicts next plan $\mathcal{P}$ given $(P_{{plan}}, x, y_{\le t})$
         \STATE Append $y_t$ to $y$
         \ENDWHILE
         \ENDIF
         \STATE \textbf{return} $y$
    \end{algorithmic}
\label{alg:inference}
\end{algorithm}

Figure~\ref{figure:model} and Algorithm~\ref{alg:inference} presents an overview of RPG at inference. 
During the inference phase, the RPG framework enhances response quality by iteratively invoking the plan-answer capability. This approach not only provides additional knowledge to the LLM but also ensures topic consistency.
To reduce costs, the bge-reranker~\cite{bge_embedding} is employed instead of ChatGPT to select fine-grained on-topic paragraphs during the inference phase.

Specifically, for every user input $x$ and retrieved passages $\mathcal{D}_x$, the LLM $\mathcal{M}$ first determines whether additional information is needed. If $\mathcal{M}$ generates $\texttt{<\textsc{not\_need\_extra\_info}>}$, the LLM  $\mathcal{M}$ predicts the output $y$ directly using prompt $P_{ans}$ and input $x$. In other cases, relevant information about plan tokens $\mathcal{P}$ in retrieved passages is selected as fine-grained paragraphs to supplement $\mathcal{M}$ with external knowledge. Furthermore, the LLM $\mathcal{M}$, using answer prompt $P_{ans}$, then incorporates fine-grained paragraphs into the generation of the next output segment $y_t$. This segment $y_t$ is subsequently appended to $y$. The plan-answer process is repeated until the $\texttt{<\textsc{EOS}>}$ token is generated, at which point $y$ is output as the final answer.

%% file: content/experiments.tex
\begin{table*}[t]
    \centering
    \begin{tabular*}{0.7\textwidth}{@{\extracolsep{\fill}}lcccc@{}}
    \toprule[1pt]
    \multirow{2}{*}{LLMs} & \multicolumn{2}{c}{ASQA} & \multicolumn{2}{c}{ELI5} \\
    & (rougeLsum) & (mauve) & (rougeLsum) & (mauve) \\
    \midrule
    \multicolumn{5}{c}{\textit{SOTA LLMs}} \\
    ChatGPT & 36.2 & 68.8 & \textcolor{gray}{\textbf{22.8}} & 32.6 \\
    Ret-ChatGPT & \textcolor{gray}{\textbf{39.9}} & 79.7 & 20.6 & \textcolor{gray}{\textbf{57.2}} \\
    \midrule
    \multicolumn{5}{c}{\textit{Baselines without Retrieval}} \\
    Llama2\textsubscript{7B} & 15.3 & 19.0 & 18.3 & 32.4 \\
    Alpaca\textsubscript{7B} & 29.4 & 61.7 & - & - \\
    Llama2\textsubscript{13B} & 12.4 & 16.0 & 18.2 & 41.4 \\
    Alpaca\textsubscript{13B} & 32.0 & 70.6 & - & - \\
    \midrule
    \multicolumn{5}{c}{\textit{Baselines with Retrieval}} \\
    Llama2\textsubscript{7B} & 22.1 & 32.0 & 18.6 & 35.3 \\
    Alpaca\textsubscript{7B} & 33.3 & 57.9 & - & - \\
    Llama2-FT\textsubscript{7B} & 35.8 & 51.2 & - & - \\
    Llama2\textsubscript{13B} & 20.5 & 24.7 & 18.6 & 42.3 \\
    Alpaca\textsubscript{13B} & 36.7 & 56.6 & - & - \\
    SuRe\textsubscript{7B}	& 35.8	& 76.2	& - & - \\
    RECOMP\textsubscript{abstractive, 7B} & 36.5 & 76.0 & - & - \\
    Self-RAG\textsubscript{7B} & 35.7 & 74.3 & 17.9 & 35.6 \\
    Self-RAG\textsubscript{13B} & 37.0 & 71.6 & - & - \\
    \midrule
    RPG\textsubscript{7B} & \textbf{37.6} & \textbf{84.4} & \textbf{19.1} & \textbf{46.4} \\
    \bottomrule[1pt]
    \end{tabular*}
    \caption{Experimental results on long-form tasks. Bold numbers indicate the best performance, except for ChatGPT.}
    \label{tab:long-form}
\end{table*}

\subsection{Experiment Setup}
To validate the effectiveness of our Plan-Retrieve-Generation framework, 
we conduct in-depth experiments on 5 carefully selected knowledge-intensive tasks. Aligning with the previous work~\cite{asai2023self}, we conduct zero-shot evaluations and utilize metrics focused on assessing the correctness, factuality, and fluency of outputs.

\subsubsection{Tasks and Datasets}

\paragraph{Long-form generation tasks.} The long-form QA tasks aim to generate comprehensive answers to questions seeking complex information, which is a primary application scenario for our model. And evaluations of these tasks can serve as evidence to the frameworks' capability of generating on-topic and comprehensive answers. We utilize ASQA~\cite{stelmakh-etal-2022-asqa} and ELI5~\cite{fan-etal-2019-eli5} as our testbed, where inputs are ambiguous questions with multiple interpretations, and outputs are expected to address them comprehensively. 
Following Self-RAG~\cite{asai2023self} and ALCE~\cite{gao-etal-2023-enabling}, we use ROUGE- L~\cite{lin2004rouge} and MAUVE~\cite{pillutla2021mauve} for correctness and fluency evaluations.

\paragraph{Multi-hop generation tasks.} A multi-hop QA task aims to test reasoning and inference skills by requiring a model to read multiple paragraphs and answer a given question. We use the 2WikiMultiHopQA~\cite{ho2020constructing} dataset and adopt the F1 score as the metric.

\paragraph{Short-form generation tasks.} The short-form QA tasks aim to generate precise answers for users, which evaluate the model's ability to effectively leverage retrieved information to response precisely. We use two open-domain QA datasets, PopQA~\cite{mallen2022not} and PubHealth~\cite{zhang2023interpretable}, where models need to answer arbitrary questions about factual knowledge. We process these two datasets following~\cite{asai2023self}.

\subsubsection{Baselines}  
Our training dataset is derived from Self-RAG and HotpotQA, where each sample is divided into planning and answering segments using ChatGPT. To ensure a fair comparison, we select baseline models that are fundamentally consistent with Self-RAG and categorize them into three major groups.

\paragraph{Baselines without retrieval.}
To explore the specific impact of external knowledge on model performance, several retrieval-free baselines are established. 
We evaluate the open-source models Llama2\textsubscript{7B, 13B} and Alpaca\textsubscript{7B, 13B}~\cite{touvron2023llama}, which have shown outstanding performance on various tasks. 

\paragraph{Baselines with retrieval.} 
We further set up baseline models with retrieval, covering the standard RAG systems. The standard RAG generates content by merging the query and retrieved documents into the input.  
We also compare the full-parameter fine-tuned version of Llama2-FT based on Self-RAG train data. Additionally, We included baseline methods related to knowledge extraction, such as SuRe~\cite{kim2024sure} and RECOMP~\cite{xu2023recomp}. These methods help LLMs make more accurate predictions by summarizing and extracting key information from retrieved text paragraphs once. However, they do not incorporate an iterative refinement process to fully utilize the knowledge. And we evaluate the Self-RAG model, which enhances the standard RAG by introducing dynamic retrieval and reflection tokens.

\paragraph{ChatGPT-Based baselines.}Lastly, we conduct a comparison with the SOTA in the field of LLMs: ChatGPT and Ret-ChatGPT (ChatGPT with retrieval passage). As a leading LLM, ChatGPT has demonstrated exceptional performance across multiple domains, providing a strong comparative benchmark for our model.

\begin{table*}[t]
    \centering
    \begin{tabular*}{0.7\textwidth}{@{\extracolsep{\fill}}lcccc@{}}
    \toprule[1pt]
    \multirow{3}{*}{LLMs} & \multicolumn{1}{c}{Multi-Hop} & \multicolumn{2}{c}{Short-form} \\
    & 2WikiMultiHopQA & PopQA & PubHealth \\
    & (F1) & (acc) & (acc) \\
    \midrule
    \multicolumn{5}{c}{\textit{SOTA LLMs}} \\

    ChatGPT & 24.8 & 29.3 & 70.1 \\
    Ret-ChatGPT & 32.8 & 50.8 & 54.7 \\
    SuRe \textsubscript{GPT} & \textcolor{gray}{\textbf{38.1}} & - & - \\
    \midrule
    \multicolumn{5}{c}{\textit{Baselines without Retrieval}} \\
    
    Llama2\textsubscript{7B} & 18.9 & 14.7 & 34.2 \\
    Alpaca\textsubscript{7B} & - & 23.6 & 49.8 \\
    Llama2\textsubscript{13B} & 20.2 & 14.7 & 29.4 \\
    Alpaca\textsubscript{13B} & - & 24.4 & 55.5 \\
    \midrule
    \multicolumn{5}{c}{\textit{Baselines with Retrieval}} \\

    Llama2\textsubscript{7B} & 21.0 & 38.2 & 30.0 \\
    Alpaca\textsubscript{7B} & - & 46.7 & 40.2 \\
    Llama2-FT\textsubscript{7B} & - & 48.7 & 64.3 \\
    Llama2\textsubscript{13B} & 31.2 & 45.7 & 30.2 \\
    Alpaca\textsubscript{13B} & - & 46.1 & 51.1 \\
    SuRe \textsubscript{Llama2 7B} & 20.6 & - & - \\
    RECOMP\textsubscript{abstractive} & 32.4 & - & - \\

    Self-RAG\textsubscript{7B} & 25.1 & 54.9 & 72.4 \\
    Self-RAG\textsubscript{13B} & - & 55.8 & \textbf{74.5} \\
    \midrule 
    RPG\textsubscript{7B} & \textbf{33.6} & \textbf{56.0} & 73.4 \\
    \bottomrule[1pt]
    \end{tabular*}
    \caption{Experimental results on Multi-Hop and Short-form generation tasks. Bold numbers indicate the best performance, except for ChatGPT.}
    \label{tab:multi-short-form}
\end{table*}

\subsubsection{Implementation Details}

\paragraph{Training.} As mentioned before, our training data is reconstructed from the Self-RAG dataset. We adopt Llama2\textsubscript{7B} as our foundational LLM, and use the prompt tuning implementation of the Huggingface PEFT ~\cite{peft} library to fine-tune LLama2\textsubscript{7B} on 4 Nvidia A6000 GPUs.

\paragraph{Inference.} During inference, the plan and answer stages alternate, using a simple greedy decoding strategy. The plan phase has a token limit of 30, and the answering phase is 100. For short-form QA, the model only completes one plan-answer cycle. For long-form and multi-hop QA tasks, the model alternates between planning and answering until it generates a termination symbol or reaches the operation limit (3 in this paper). In multi-hop QA, a special "[Combine]" symbol indicates that the model will summarize the previous content to produce a concise answer. For the retriever model, we use Contriever-MS MARCO for PopQA, PubHealth, and ASQA datasets, and BM25 for the 2WikiMultiHop datasets, aligning with all baselines.

\subsection{Experiment Results}

\paragraph{Long-form generation.} Our model demonstrates brilliant performance in the domain of long-form generation, which is the primary application scenario for our model. 
As Table \ref{tab:long-form} displayed, the experimental results demonstrate that our model has achieved a significant improvement in long-form generation performance with only a slight tuning of 0.3 billion parameters. Notably, our model outperforms the prior SOTA Self-RAG. 
% Specifically, on the ASQA dataset, our model outperforms Self-RAG by 2 points on the ROUGE metric, which measures the correctness and comprehensiveness of long-form generation. Additionally, on MAUVE, a newly introduced metric for evaluating the fluency and coherence of model-generated text, our model significantly outperforms the Self-RAG model by more than 10 points. 
Compared to the knowledge-extraction baseline, both SuRe and RECOMP-abstractive extract key information by compressing and rewriting the retrieved content once. In contrast, our method is iterative, extracting key information based on explicit planning until the content is sufficient. RPG outperforms these methods, underscoring the significance of iterative planning. Single-step extraction and answering can overlook intermediate information, while iterative planning with on-demand retrieval is more effective.
Even when compared to the current SOTA model, ChatGPT with retrieved knowledge, our model achieves comparable results. Similar findings are also observed in the ELI5 dataset.

These results underscore our model's strong capabilities in long-form generation tasks, demonstrating the comprehensiveness and relevance of our model's responses. The iterative alternation between the planning and answering phases ensures that the generated text remains on-topic. Our approach not only enhances fluency but also maintains factual accuracy, further highlighting the superiority of our method.

\paragraph{Multi-hop generation.} For multi-hop generation tasks, the model needs to integrate all generated information to provide a concise answer.
Experimental results in Table \ref{tab:multi-short-form} indicate that our RPG framework outperforms other Llama-based baseline models, demonstrating the benefit of pre-planning and utilizing fine-grained evidence for reasoning. While the GPT-based SuRe~\cite{kim2024sure} model performs better than ours, the Llama-based SuRe model performs poorly due to its dependence on rewriting retrieved content, a process reliant on LLM's capabilities. In contrast, our model avoids this rewriting process and still achieves good performance on multi-hop datasets.

\paragraph{Short-form generation.} Although short-form generation is not the primary application scenario of our model, we still demonstrate its performance in this context to prove its versatility and applicability. In some short-form generation tasks, especially on the Pub dataset, we find that retrieved content is not always effective. 
In fact, retrieval-augmented ChatGPT often underperforms compared to its non-retrieval-augmented counterpart due to the incorporation of irrelevant information.
By focusing on the relevance of retrieved content and excluding irrelevant details, our model shows progress in various short-form generation tasks.

% \begin{table}[t]
% \resizebox{0.5\textwidth}{!}{
% \begin{tabular}{cccc}
% \toprule[1pt] \begin{tabular}{c} 
% Variations
% \end{tabular} & \begin{tabular}{c} 
% Pub \\
% (acc)
% \end{tabular} &  \begin{tabular}{c} 
% 2Wiki \\
% (F1)
% \end{tabular} &  \begin{tabular}{c} 
% ASQA \\
% (rg)
% \end{tabular} \\
% \hline RPG & \textbf{73.4} & \textbf{33.6} & \textbf{37.6} \\
% \hline 
% \multicolumn{4}{c}{\textit{Training}} \\
%  No Plan & 69.1   & 27.4 & 32.0 \\
% No Multi-task Learning & 70.1 & 23.1 & 34.1 \\
% \hline \multicolumn{4}{c}{\textit{Inference}} \\
%  No Retrieval & 72.3 & 27.0 & 32.4 \\
%  No Paragraph Selection  & 72.3 & 30.2 & 34.8 \\
% \bottomrule[1pt]
% \end{tabular}
% }
% \caption{Ablations in training and inference.  }
% \label{tab:ablation}
% \end{table}

\subsection{Ablations}

As shown in Table ~\ref{tab:ablation}, we conduct a comprehensive ablation study on the RPG framework to clarify which factors play a decisive role in the training and inference processes. 

\paragraph{Training Phase.}
We investigate the impact of removing the planning phase on model performance. By eliminating all plan texts from the training dataset and using prompt tuning to train the model with only answer texts, we observe a significant drop in performance across all three tasks. In long-form generation, the absence of planning caused the model to deviate from the topic. In short-form generation, unscreened retrieved texts were not always beneficial. Thus, the planning phase is crucial for maintaining the relevance of generated content.

Furthermore, we investigate the differences between fine-tuning a model with uniform learnable prompt tokens for both plans and answers versus using distinct tokens for each. Results show that uniform tokens diminished performance in both long-term and short-term generation tasks, suggesting that planning and answering function as separate tasks. Thus, it is more appropriate to use multi-task learning to train LLMs for both planning and answering capabilities.

\begin{figure}[t]
\centering
\includegraphics[width=0.49\textwidth]{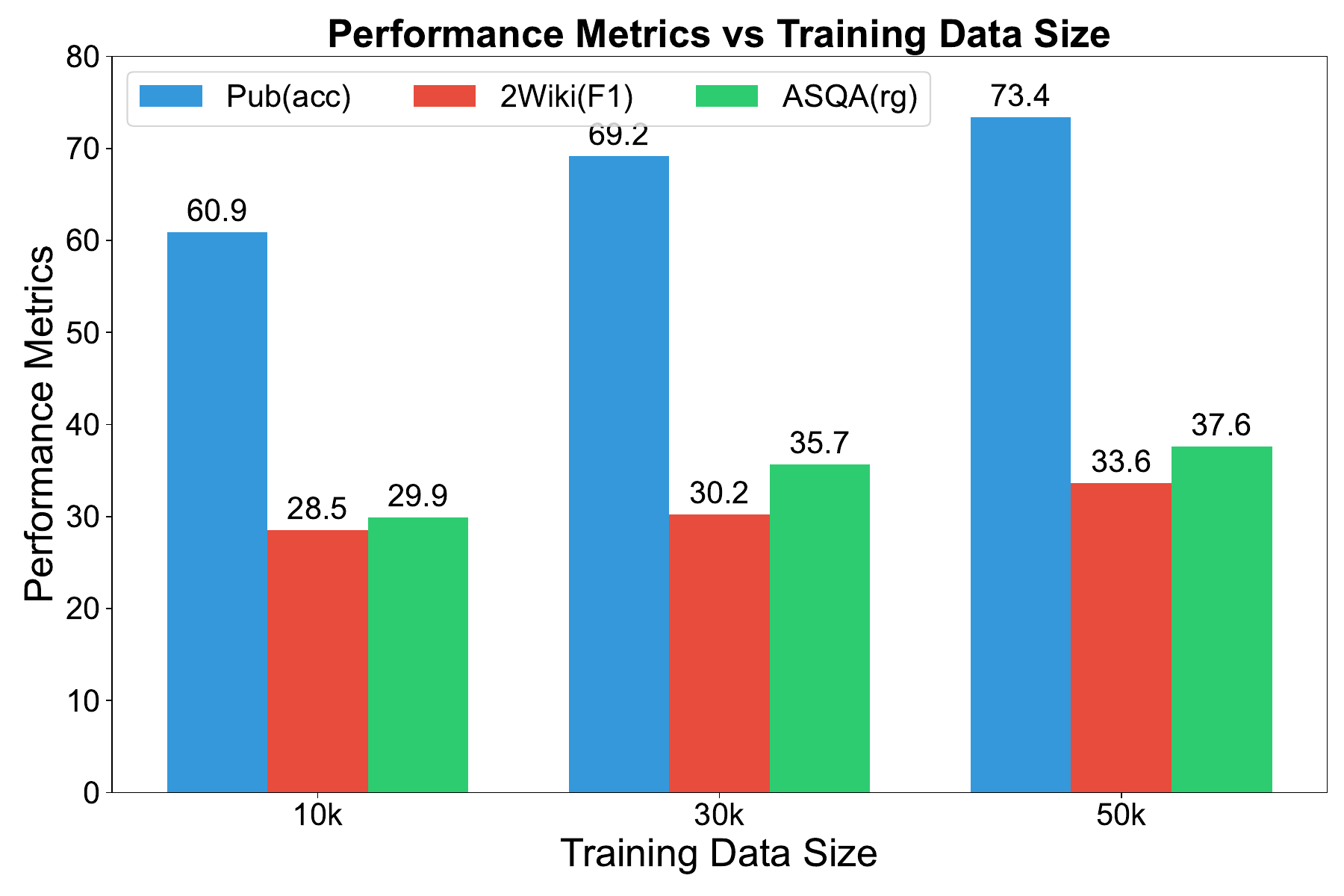}
\caption{Training scale analysis.}
\label{figure:training_scale}
\end{figure}

Additionally, we study the model's performance with varying scales of training datasets as Figure ~\ref{figure:training_scale} displayed. The results show that performance gradually improves as the dataset size increases. We believe further expanding the training data will continue to enhance the model's performance.

\begin{table}[t]
\centering
\renewcommand{\arraystretch}{1.1} % 调整行间距
\resizebox{0.5\textwidth}{!}{
\begin{tabular}{cccc}
\toprule[1pt]
    
\multirow{2}{*}{Variations}  & Pub & 2Wiki & ASQA \\
    & (acc) & (F1) & (rougeLsum) \\

\midrule
RPG & \textbf{73.4} & \textbf{33.6} & \textbf{37.6} \\
\midrule
\multicolumn{4}{c}{\textit{Training}} \\

No Plan & 69.1 & 27.4 & 32.0 \\

\begin{tabular}[c]{@{}c@{}}No Multi-task  Learning\end{tabular} & 70.1 & 23.1 & 34.1 \\

\midrule
\multicolumn{4}{c}{\textit{Inference}} \\

No Retrieval & 72.3 & 27.0 & 32.4 \\

\begin{tabular}[c]{@{}c@{}}No Para Selection\end{tabular} & 72.3 & 30.2 & 34.8 \\

Plan globally	& - & -	& 33.1 \\

\bottomrule[1pt]
\end{tabular}
}
\caption{Ablations in training and inference.}
\label{tab:ablation}
\end{table}

\paragraph{Inference Phase.}
In the inference phase, we assess the impact of retrieval on model performance. Results show that retrieval is crucial for long-form generation tasks, which require comprehensive answers. Without retrieval, generating complete answers is significantly more challenging. Conversely, for short-term generation tasks, retrieval has a minor impact, since these tasks may typically do not require extensive knowledge.

Additionally, we examine the effects on model performance when using retrieved passages directly. The results show a significant decline in performance across all tasks, highlighting the detrimental impact of off-topic paragraphs on the quality of generated outputs.

% In this section, we discuss the performance of the RPG method under different settings. 

Eventually, we compared global planning and local planning methods. Since RPG involves iteratively generated local plans, we employed GPT-3.5 model to globally annotate all plans for the question. The results in the table shows iterative planning (RPG) outperforms global planning (Plan globally) in long-form generation, indicating that iterative planning more effectively generates high-quality answers for the current tasks.

% Additionally, we compared the impact of different fine-grained paragraph extractors on the RPG method, utilizing both GPT-3.5 and BGE-reranker for comparison. Results shown in Table~\ref{tab:extractor} indicates that the performance of both methods is very similar.

%% file: content/conclusion.tex
In this paper, we propose a Retrieve-Plan-Generation (RPG) framework, which integrates an explicit plan stage into the lengthy generation process. By generating plan tokens, the model is guided to selectively utilize retrieved paragraphs. The iterative alternation between plan and answer stages ensures that the generated content remains relevant to the topic. To implement this framework, we adopt an efficient multi-task fine-tuning method that equips existing models with both planning and answering capabilities. Experimental results demonstrate that RPG outperforms state-of-the-art models across five tasks, validating the effectiveness of our approach.

%% file: content/limitations.tex
Due to computational resource constraints, we only present the specific implementation of the RPG framework under the Llama2\textsubscript{7B}, without exploring further experiments on larger models, such as Llama2\textsubscript{13B}, Llama2\textsubscript{70B}. Additionally, due to the API costs associated with accessing ChatGPT, we conducted experiments solely on a 50k reconstructed dataset, without collecting and analyzing more extensive data to provide more experimental results on larger datasets.

\section{Acknowledgment}
This work was supported by the grants from National Natural Science Foundation of China (No.62222213, 62072423).

%% file: content/appendix.tex
\section{More PRG Implementation Details}
\paragraph{Training} As previously mentioned, our training data is structured based on the Self-RAG dataset.

During the training phase, we utilize the Llama2\textsubscript{7B} as our foundational language model. For the retriever model, we have selected the readily accessible Contriever-MS MARCO for the PopQA, Pub, and ASQA datasets, and the BM25 algorithm for the 2Wiki datasets, aligning with the baselines

\begin{table*}[t]
    \centering
    \begin{tabular}{l|l|l|r}
        \hline
        \textbf{Dataset name} & \textbf{Category} & \textbf{Data source} & \textbf{\# of instances} \\
        \hline
        ShareGPT & Instruction-following & Open-Instruct & 13,095 \\
        Natural Questions & Knowledge-intensive & KILT & 15,226 \\
        FEVER & Knowledge-intensive & KILT & 9,665 \\
        OpenBoookQA & Knowledge-intensive & HF Dataset & 4,699\\
        Arc-Easy & Knowledge-intensive & HF Dataset & 1847 \\
        ASQA & Knowledge-intensive & ASQA & 3,564 \\
        HotpotQA & Knowledge-intensive & HotpotQA & 3830 \\
        \hline
    \end{tabular}
    \caption{Dataset statistics}
    \label{tab:datasets}
\end{table*} 

\begin{table*}[t]
    \centering
    \begin{tabular}{l|l|r|r}
        \hline
        \textbf{Dataset name} &  \textbf{Data source} & \textbf{Avg. \# of Plan} & \textbf{\% of Plan=}\texttt{True}\\
        \hline
        ShareGPT &  Open-Instruct & 3.939 & 70.3\\
        Natural Questions  & KILT & 0.877 & 87.7 \\
        FEVER  & KILT & 0.634 & 63.4 \\
        OpenBoookQA  & HF Dataset & 0.023 & 2.3 \\
        Arc-Easy  & HF Dataset & 0.108 & 10.8 \\
        ASQA  & ASQA & 1.916 & 91.6 \\
        HotpotQA  & HotpotQA & 1.338 & 77.8 \\
        \hline
    \end{tabular}
    \caption{Detailed plan statistics of dataset}
    \label{tab:plan info}
\end{table*}

\begin{table*}[t]
\centering
\begin{tabular}{l|c|c|c}
\hline
\textbf{Comparison} & \textbf{Better} & \textbf{Equal} & \textbf{Worse} \\ \hline
RPG vs. Self-RAG (Comprehensiveness) & 44\% & 16\% & 40\% \\ \hline
RPG vs. Self-RAG (Correctness) & 34\% & 32\% & 34\% \\ \hline
RPG vs. RPG without Para Selection (Comprehensiveness) & 55\% & 9\% & 36\% \\ \hline
RPG vs. RPG without Para Selection (Correctness) & 41\% & 23\% & 36\% \\ \hline
\end{tabular}%
\caption{Subjective Evaluation Results}
\label{tab:subjective eval}
\end{table*}

\paragraph{Inference} During the inference process, the planning and answering stages alternate, and we have employed a simple greedy decoding strategy for both. In the planning phase, we set a maximum token generation limit of 30, while in the answer phase, it is 100. As for the retrieved documents, by default, we use the top five documents ranked by Contriever-MS MARCO ~\cite{izacard2021unsupervised}; For ASQA, we utilized the top five documents selected by the authors from GTR-XXL~\cite{ni2021large}, which is done to ensure a fair comparison among all baseline models. 
In short-form QA, the model executes a single planning and answering cycle. Conversely, in long-form QA tasks, the model alternates between planning and answering multiple times until it generates a termination symbol or reaches the limit of operations(3 in this paper). Multi-hop QA follows a similar approach to long-form QA. However, there is a minor difference: as the generation process nears completion, our model generates a special "[Combine]" symbol. This indicates that the model will then summarize the previously generated content and ultimately produce a concise answer to the original question.

\section{Statistical information of the Dataset}
\label{appendix:dataset statistical}
In this section, we provide a detailed discussion of the statistical information and relevant details of the dataset. The statistical information of the experimental data is presented in Table~\ref{tab:datasets}, with additional statistics on the dataset's Plan information shown in Table~\ref{tab:plan info}.

\section{Additional Analysis for the Plan Module}
As a crucial component of the RPG framework, the Plan module plays an essential role in guiding model generation and extracting fine-grained evidence. We further discuss the impact of the Plan module under different settings on model generation, providing intuitive explanations and analysis using examples.

\textbf{\textit{Plan globally vs locally.}}
Plan Tokens can be generated both globally and locally. We use ChatGPT to create an initial global plan. Our work, however, emphasizes an iterative plan-answer process, viewed as a local plan. A case analysis in Table~\ref{tab:global local case} shows that RPG produces more comprehensive plans, while ChatGPT focuses on a single question aspect, leading to less detailed responses, which also demonstrates the importance of the iterative process. 

\begin{table*}[t]
    \centering
    \resizebox{\textwidth}{!}{
    \begin{tabular}{p{17cm}}
        \hline
        \textbf{Question}: Which is the most popular sport in Australia?\\
        \hline
        \textbf{Plan globally with ChatGPT}: \textcolor[rgb]{0.9,0,0.1}{<plan\_start>Identify popular sports in Australia<plan\_end>}\textcolor[rgb]{0,0.7,0.3}{Fine-grained evidence}\textcolor[rgb]{1,0.65,0}{<answer\_start>Australian Rules Football (AFL) is considered the most popular sport in Australia,<answer\_end>}\textcolor[rgb]{0.9,0,0.1}{<plan\_start>Participation rates and viewership<plan\_end>}\textcolor[rgb]{0,0.7,0.3}{Fine-grained evidence}\textcolor[rgb]{1,0.65,0}{<answer\_start>with the highest participation rates and viewership, particularly in Victoria, South Australia, Western Australia, and Tasmania.<answer\_end>}\\
        \hline
        \textbf{Ours}: There are a lot of popular sports in Australia.\textcolor[rgb]{0.9,0,0.1}{<plan\_start>most popular sport in Australia<plan\_end>}\textcolor[rgb]{0,0.7,0.3}{Fine-grained evidence}\textcolor[rgb]{1,0.65,0}{<answer\_start>As a whole, the most popular sport to play in Australia is Australian Rules Football.<answer\_end>}\textcolor[rgb]{0.9,0,0.1}{<plan\_start>Most popular spectator sport in different regions of Australia<plan\_end>}\textcolor[rgb]{0,0.7,0.3}{Fine-grained evidence}\textcolor[rgb]{1,0.65,0}{<answer\_start>Historically, rugby league and rugby union football codes have been more popular than Australian rules football in New South Wales and Queensland, whereas Australian rules football has been more popular in Victoria, Tasmania, South Australia, North Territory and Western Australia.<answer\_end>}\textcolor[rgb]{0.9,0,0.1}{<plan\_start>Most popular spectator sport in different genders of Australia<plan\_end>}\textcolor[rgb]{0,0.7,0.3}{Fine-grained evidence}\textcolor[rgb]{1,0.65,0}{<answer\_start>For adult women in Australia, the number one sport activity they participate in is walking with 30\% having done this in 2009 and 2010.For men, the most popular sport activity was also walking with a participation rate of 15.6\%.<answer\_end>}\\
        \hline
    \end{tabular}
    }
    \caption{An  example for global/local plan}
    \label{tab:global local case}
\end{table*}

\textbf{\textit{More plans, better generation?}}
Although the plan can enhance the model's understanding of questions, allowing it to focus on relevant fine-grained information and improve generation quality, an excess of planning can lead to a decline in results. For better illustration, we forced RPG to generate one more plan, a corresponding example is shown in Table~\ref{tab:one more plan}. It is clear that forcing the model to continue generating when it has already produced sufficient content results in repetitive plans and answers, leading to a decline in generation quality.

\begin{table*}[t]
    \centering
    \resizebox{\textwidth}{!}{
    \begin{tabular}{p{17cm}}
        \hline
        \textbf{Question}: When was the first Toyota Land Cruiser made?\\
        \hline
        \textbf{Ours}: \textcolor[rgb]{0.9,0,0.1}{<plan\_start>toyota land cruiser<plan\_end>}\textcolor[rgb]{0,0.7,0.3}{Fine-grained evidence}\textcolor[rgb]{1,0.65,0}{<answer\_start>The Land Cruiser is a series of four-wheel drive vehicles produced by the Japanese automobile manufacturer Toyota.<answer\_end>}\textcolor[rgb]{0.9,0,0.1}{<plan\_start>first Toyota Land Cruiser made in Japan<plan\_end>}\textcolor[rgb]{0,0.7,0.3}{Fine-grained evidence}\textcolor[rgb]{1,0.65,0}{<answer\_start>The first Toyota Land Cruiser was made in Japan in 1951. <answer\_end>}\\
        \hline
        \textbf{Forced continuation}: \textcolor[rgb]{0.9,0,0.1}{<plan\_start>toyota land cruiser<plan\_end>}\textcolor[rgb]{0,0.7,0.3}{Fine-grained evidence}\textcolor[rgb]{1,0.65,0}{<answer\_start>The Land Cruiser is a series of four-wheel drive vehicles produced by the Japanese automobile manufacturer Toyota.<answer\_end>}\textcolor[rgb]{0.9,0,0.1}{<plan\_start>first Toyota Land Cruiser made in Japan<plan\_end>}\textcolor[rgb]{0,0.7,0.3}{Fine-grained evidence}\textcolor[rgb]{1,0.65,0}{<answer\_start>The first Toyota Land Cruiser was made in Japan in 1951. <answer\_end>}\textcolor[rgb]{0.9,0,0.1}{<plan\_start>first Toyota Land Cruiser made in Japan<plan\_end>}\textcolor[rgb]{0,0.7,0.3}{Fine-grained evidence}\textcolor[rgb]{1,0.65,0}{<answer\_start>The Land Cruiser is a series of four-wheel drive vehicles produced by the Japanese automobile manufacturer Toyota.<answer\_end>}\\
        \hline
    \end{tabular}
    }
    \caption{An  example for forcing plan}
    \label{tab:one more plan}
\end{table*}

\section{Subjective Evaluation of Generation}
We further explore the subjective evaluation results of the content generated by the RPG method. Given that existing evaluation metrics are already sufficient for short-form and multi-hop QA tasks, we focus primarily on the subjective assessment for long-form QA tasks.

To save costs, we adopted an evaluation method consistent with SelfRAG. For the ASQA dataset, we extracted 50 different questions and responses from two different models, and evaluated them using two human annotators and GPT-4o from the following two aspects:

\begin{enumerate}
    \item \textbf{Comprehensiveness}: Which answer contains more correct content, considers more knowledge, and provides a more comprehensive response to the question.

    \item \textbf{Correctness}: Which answer is more accurate in relation to the question.
\end{enumerate}

% The evaluation involves determining which answer performs better in terms of comprehensiveness and correctness. 
We collected annotation data from GPT-4o and humans and found a high agreement of 94\%. We consider GPT-4o’s annotations to be reliable. Then we conducted a comprehensive experiment with GPT-4o on ASQA, comparing SelfRAG and RPG, as well as RPG with and without the plan phase for fine-grained paragraph selection. The results are shown in Table~\ref{tab:subjective eval}.

% \begin{table*}[h]
% \centering
% \begin{tabular}{l|c|c|c}
% \hline
% \textbf{Comparison} & \textbf{Better} & \textbf{Equal} & \textbf{Worse} \\ \hline
% RPG vs. Self-RAG (Comprehensiveness) & 44\% & 16\% & 40\% \\ \hline
% RPG vs. Self-RAG (Correctness) & 34\% & 32\% & 34\% \\ \hline
% RPG vs. RPG without Paragraph Selection (Comprehensiveness) & 55\% & 9\% & 36\% \\ \hline
% RPG vs. RPG without Paragraph Selection (Correctness) & 41\% & 23\% & 36\% \\ \hline
% \end{tabular}%
% \caption{Subjective Evaluation Results}
% \label{tab:subjective eval}
% \end{table*}

% Comparison	Better	Equal	Worse
% RPG vs. Self-RAG (Comprehensiveness)	44%	16%	40%
% RPG vs. Self-RAG (Correctness)	34%	32%	34%
% RPG vs. RPG without Paragraph Selection (Comprehensiveness)	55%	9%	36%
% RPG vs. RPG without Paragraph Selection (Correctness)	41%	23%	36%
The experimental results are consistent with the quantitative results. Our model performs better than SelfRAG, and the plan phase generates plans that effectively help us focus on key information, resulting in more comprehensive and factually accurate information.

% \section{RPG Method Performance Analysis}
% In this section, we discuss the performance of the RPG method under different settings. First, we compared global planning and local planning methods. Since RPG involves iteratively generated local plans, we employed GPT-3.5 model to globally annotate all plans for the question. The results of the two plan generation approaches on ASQA are presented in Table~\ref{tab:global vs local}. These results show that iterative planning (RPG-iter) outperforms global planning (RPG-global) in both Rouge-L and MAUVE metrics, indicating that iterative planning more effectively generates high-quality answers for the current tasks.

% Additionally, we compared the impact of different fine-grained paragraph extractors on the RPG method, utilizing both GPT-3.5 and BGE-reranker for comparison. Results shown in Table~\ref{tab:extractor} indicates that the performance of both methods is very similar.

% \begin{table*}[t]
%     \centering
%     \begin{tabular}{c|cc}
%     \hline
%     \textbf{Model}	& \textbf{ROUGE-L}	& \textbf{MAUVE} \\ \hline
% RPG-global	& 33.1	& 73.8 \\ \hline
% RPG-iter	& 37.6	& 84.4 \\ \hline
%     \end{tabular}
%     \caption{Comparision of RPG global and local}
%     \label{tab:global vs local}
% \end{table*}

% \begin{table*}[t]
%     \centering
%     \begin{tabular}{c|cc}
%     \hline
%     \textbf{Model}	& \textbf{ROUGE-L}	& \textbf{MAUVE} \\ \hline
% GPT-3.5 &	37.4 &	83.3 \\ \hline
% bge-reranker &	37.6 &	84.4 \\ \hline
%     \end{tabular}
%     \caption{Comparision of different extractor}
%     \label{tab:extractor}
% \end{table*}

% \newpage

\section{Prompts for Dataset construction and Examples}
\label{appendix:prompts}
In this section, we provide a detailed explanation of the construction methods for each dataset. We first introduce the instructions used for dataset construction and then provide corresponding examples for each dataset. 

To construct our own dataset, we utilize \texttt{gpt-3.5-turbo-0125} to generate comprehensive annotations leveraging existing datasets and few-shot examples. Given the straightforward nature of the short-form questions, we prompt ChatGPT to summarize their statements as the Plan, which is outlined in Figure~\ref{tab:plan short}. We apply this method to Natural Questions, FEVER, OpenBoookQA, and Arc-Easy. 
ASQA consists of numerous ambiguous questions, where each problem within the annotated dataset is further divided into multiple sub-problems post artificial disambiguation. These segments address specific parts of the question, and due to their close resemblance, ChatGPT may generate very similar topics based on answer summaries. ChatGPT should identify which sub-problems the current answer corresponds to, and then summarize these sub-problems into a statement. The process is guided by prompts detailed in Table~\ref{tab:plan asqa}. As ShareGPT does not encounter many ambiguous questions, for each part of the answer, we directly prompt ChatGPT to summarize the current segment's topic based on the provided answer context as the label for the Plan Generation. Detailed information is provided in Table~\ref{tab:plan sharegpt}.
For HotpotQA, since there is sufficient evidence in the annotated data and the question only needs two jumps at most, we believe that ChatGPT is sufficient to give good planning based on the question and answer. The instructions are shown in Table~\ref{tab:plan hotpotqa}. Prompts used for fine-grained evidence selection are shown in Table~\ref{tab:evidence}. Examples of our dataset can be found in Table~\ref{tab:data short examples}, Table~\ref{tab:asqa data examples}, and Table~\ref{tab:hotpotqa data examples}.

\begin{table*}
\resizebox{\textwidth}{!}{
\begin{tcolorbox}[colback=gray!10,
                  colframe=black,
                  width=\linewidth,
                  arc=1mm, auto outer arc,
                  boxrule=0.5pt,
                  title=Instructions for Plan Generation of Short-form QA,
                 ]
\textit{\textbf{Plan Generation:}}
\\
\textbf{Instructions:}
\\
Extract the body of the statement from the question into a Plan token. The plan token should be like [Plan: XX].

Input: which company Javed Afridi is best known as CEO? 

Output: [Plan: Javed Afridi best known company]. 

Input: \textit{a question}

Output:

\end{tcolorbox}
}
\caption{Instructions for Constructing Plan Generation Datasets of Short-form QA}
\label{tab:plan short}
\end{table*}

\begin{table*}
\resizebox{\textwidth}{!}{
\begin{tcolorbox}[colback=gray!10,
                  colframe=black,
                  width=1\textwidth,
                  arc=1mm, auto outer arc,
                  boxrule=0.5pt,
                  title=Instructions for Plan Generation of ASQA,
                 ]
\textit{\textbf{Plan Generation:}}
\\
\textbf{Instructions:}
\\
Given several short qa-pairs and a sentence, you need to decide which qa-pair is this sentence relevant to. Always cite for any factual claim. When citing several search results, use \texttt{[1]}\texttt{[2]}\texttt{[3]}. If multiple qa-pairs support the sentence, only cite a minimum sufficient subset of the qa-pairs. \\
QA-Pairs: \\
\texttt{[0]} Where Haier Pakistan is located? Pakistan. \\
\texttt{[1]} When was Haier Pakistan established? 2000.\\
Sentence: \\
Established in 2000, it is a subsidiary of the Chinese multinational group Haier.\\
Out: \texttt{[1]}\\
QA-Pairs: \\
\texttt{[0]} When does episode 42 of bunk'd come out? May 24, 2017. \\
\texttt{[1]} When does episode 41 of bunk'd come out?? April 28, 2017.\\
\texttt{[2]} When does episode 40 of bunk'd come out? April 21, 2017.\\
Sentence: \\
The new bunk'd episode 41 comes out on April 21, 2017, episode 42 comes out on April 28, 2017 and episode 42 is due to come out on May 24, 2017.\\
Out: \texttt{[0]}\texttt{[1]}\texttt{[2]}\\
QA-Pairs: \\
your qa-pairs\\
Sentence:\\
your sentence\\
Out:
\tcblower
Given a number of questions, you need to summarize them as concisely and accurately as possible into one question, avoiding missing information about each question. You don't have to answer these questions. \\
Questions: \\

\texttt{[0]} first question\\
\texttt{[1]} second question\\
\dots
\\
Output:\\
\end{tcolorbox}
}
\caption{Instructions for Plan Generation of ASQA}
\label{tab:plan asqa}
\end{table*}

\begin{table*}
\resizebox{\textwidth}{!}{
\begin{tcolorbox}[colback=gray!10,
                  colframe=black,
                  width=1\textwidth,
                  arc=1mm, auto outer arc,
                  boxrule=0.5pt,
                  title=Instructions for Plan Generation of ShareGPT]
\textit{\textbf{Plan Generation:}}
\\
\textbf{Instructions:}
\\
Generate appropriate Plan token in the following format: [Plan: xx], for each [Plan] based on relevant context. Be sure always generate a Plan Token for each [Plan] in order, Keep the details to be as different as possible from other Plan tokens. Do not generate a Plan Token where there is no [Plan]. \\
Input: AB is famous for his work in Parkistan Haier.[Plan] Established in 2000, it is a subsidiary of the Chinese multinational group Haier. \\
Output:AB is famous for his work in Parkistan Haier.[Plan: Parkistan Haier establish time] Established in 2000, it is a subsidiary of the Chinese multinational group Haier. \\
Input: answer segment\\
Output:

\end{tcolorbox}
}
\caption{Instructions for Plan Generation for each answer segment of ShareGPT}
\label{tab:plan sharegpt}
\end{table*}

\begin{table*}
\resizebox{\textwidth}{!}{
\begin{tcolorbox}[colback=gray!10,
                  colframe=black,
                  width=1\textwidth,
                  arc=1mm, auto outer arc,
                  boxrule=0.5pt,
                  title=Instructions for Fine-grained evidence selection,
                 ]
\textit{\textbf{Fine-grained evidence selection:}}
\\
\textbf{Instructions:}
\\
Write an accurate, engaging, and concise answer for the given question answer pair using only the provided search results (some of which might be irrelevant) and cite them properly. Use an unbiased and journalistic tone. Always cite for any factual claim. When citing several search results, use \texttt{[1]}\texttt{[2]}\texttt{[3]}. If multiple documents support the sentence, only cite a minimum sufficient subset of the documents.\\
Question: When was Haier Pakistan established? \\
Answer: 2000. \\
\texttt{[0]} Haier Pakistan is a consumer electronics and home appliances company in Pakistan. \\
\texttt{[1]} Established in 2000, it is a subsidiary of the Chinese multinational group Haier.\\
\texttt{[2]} It is one of the largest companies in Pakistan's home appliances market, in terms of sales and revenues generated. \\
Out: \texttt{[1]}\\
Ouestion: question\\
Answer: answer\\
\texttt{[0]} first evidence\\
\texttt{[1]} second evidence\\
\dots\\
Out:

\end{tcolorbox}
}
\caption{Instructions for Fine-grained evidence selection for each answer segment}
\label{tab:evidence}
\end{table*}

\begin{table*}
\resizebox{\textwidth}{!}{
\begin{tcolorbox}[colback=gray!10,
                  colframe=black,
                  width=1\textwidth,
                  arc=1mm, auto outer arc,
                  boxrule=0.5pt,
                  title=Instructions for HotpotQA,
                 ]
\textit{\textbf{Plan Generation:}}
\\
\textbf{Instructions:}
\\
Given a question and corresponding short answer. Expand the short answer to an accurate, fine-grained, and concise answer with thinking steps for the given question using only the provided search results (some of which might be irrelevant) and cite them properly. During the generation, make sure the plan token in answers start with the question and work their way up logically from the answers you already have. Use an unbiased and journalistic tone. Always cite for any factual claim. Cite at most one evidence in each sentence. If multiple documents support the sentence, only cite the first one.\\
Question: In what year was the company, for which Javed Afridi is best known as CEO, established?\\
Answer: 2000. \\
Evidence:\\ 
\texttt{[0]} Established in 2000, Haier Pakistan is a subsidiary of the Chinese multinational group Haier.\\
\texttt{[1]} Javed Afridi is best known as the CEO of Haier Pakistan and owner of MG Motors Pakistan.\\
Out: [Plan: Javed Afridi best known company]Javed Afridi is best known as the CEO of  Haier Pakistan\texttt{[1]},[Plan: Haier Pakistan establish]which was established in 2000.\texttt{[0]}\\
As mentioned before, the first plan token should be generated from question [Plan: Javed Afridi best known company], considering the answer already generated and further plan the establish time of Haier Pakistan.\\
Question: Where are Steph Curry and Lebron James both from?\\
Answer: America. \\
Evidence: \\
\texttt{[0]} Stephen Curry is a professional American basketball player with the Golden State Warriors..\\
\texttt{[1]} Lebron James is a professional American basketball player with the Lakers.\\
Out: [Plan: Steph Curry from]Steph Curry is from America\texttt{[0]},[Plan: Lebron James from]Lebron James is also from America.\texttt{[1]}\\
Question: question\\
Answer: answer\\
Evidence: your evidences\\
Out:

\end{tcolorbox}
}
\caption{Instructions for HotpotQA}
\label{tab:plan hotpotqa}
\end{table*}

\begin{table*}[t]
    \centering
    \resizebox{\textwidth}{!}{
    \begin{tabular}{p{17cm}} 
        \hline
        \textbf{Input}: What harmful phenomenon can cause atmospheric warming? A: gasses with pollution B: acid rain C: nuclear war D: melting ice caps 
          (\textcolor{gray}{\textit{OpenBoookQA}})\\
        \hline
        \textbf{Output}: \textcolor[rgb]{0.9,0,0.1}{<plan\_start>Harmful phenomenon causing atmospheric warming<plan\_end>}\textcolor{lightgray}{<paragraph>Asian brown cloud past century. The report suggested that air pollution regulations can have large amplifying effects on global warming. Another major impact is on the polar ice caps. Black carbon (soot) in the Asian Brown Cloud may be reflecting sunlight and dimming Earth below but it is warming other places by absorbing incoming radiation and warming the atmosphere and whatever it touches. Black carbon is three times more effective than carbon dioxide—the most common greenhouse gas—at melting polar ice and snow. Black carbon in snow causes about three times the temperature change as carbon dioxide in the atmosphere. On snow—even at concentrations</paragraph>}\textcolor[rgb]{0,0.7,0.3}{<fparagraph>The report suggested that air pollution regulations can have large amplifying effects on global warming.</fparagraph>}\textcolor[rgb]{1,0.65,0}{<answer\_start>A<answer\_end>} \\
        \hline
        \textbf{Input}:how long does a president have secret service protection
          (\textcolor{gray}{\textit{Natural Questions}})\\
        \hline
        \textbf{Output}: \textcolor[rgb]{0.9,0,0.1}{<plan\_start>president secret service protection duration<plan\_end>}\textcolor{lightgray}{<paragraph>Former Presidents Act health insurance under the Federal Employees Health Benefits Program;  a GSA legal opinion ruled Jimmy Carter ineligible.  Former presidents were entitled from 1965 to 1996 to lifetime Secret Service protection, for themselves, spouses, and children under 16.  A 1994 statute, (), limited post-presidential protection to ten years for presidents inaugurated after January 1, 1997.  Under this statute, Bill Clinton would still be entitled to lifetime protection, and all subsequent presidents would have been entitled to ten years' protection.  On January 10, 2013, President Barack Obama signed the Former Presidents Protection Act of 2012, reinstating lifetime Secret Service protection for his</paragraph>}\textcolor[rgb]{0,0.7,0.3}{<fparagraph>Former presidents were entitled from 1965 to 1996 to lifetime Secret Service protection, for themselves, spouses, and children under 16. Under this statute, Bill Clinton would still be entitled to lifetime protection, and all subsequent presidents would have been entitled to ten years' protection. </fparagraph>}\textcolor[rgb]{1,0.65,0}{<answer\_start>lifetime<answer\_end>}\\
        \hline
        \textbf{Input}: Is the following statement correct or not? Say true if it's correct; otherwise say false.\#\# Input: Evan Goldberg is a Canadian director.
          (\textcolor{gray}{\textit{FEVER}})\\
        \hline
        \textbf{Output}: \textcolor[rgb]{0.9,0,0.1}{<plan\_start>Evan Goldberg is a Canadian director.<plan\_end>}\textcolor{lightgray}{<paragraph>Evan Goldberg Evan Goldberg Evan Goldberg (born May 11, 1982) is a Canadian screenwriter, film producer, and director. He has collaborated with his childhood friend Seth Rogen in several films, including "Superbad" (2007) (which they first conceived as teenagers), "Pineapple Express" (2008), "This Is the End" (2013) (their directorial debut), and "The Interview" (2014). Goldberg was born in Vancouver, British Columbia, to a Jewish family. He was raised in Marpole. He attended Point Grey Secondary School (where he met Rogen) and McGill University, and is married to Lisa (Yadavaia) Goldberg. Goldberg started his writing career joining the staff of "Da Ali G</paragraph>}\textcolor[rgb]{0,0.7,0.3}{<fparagraph>Evan Goldberg Evan Goldberg Evan Goldberg (born May 11, 1982) is a Canadian screenwriter, film producer, and director.</fparagraph>}\textcolor[rgb]{1,0.65,0}{<answer\_start>true<answer\_end>}\\
        \hline
        \textbf{Input}: Given four answer candidates, A, B, C and D, choose the best answer choice. \#\# Input: Darwin's theory that animal species can change over time was inspired by his research on which set of islands? A: the Philippine Islands B: the Virgin Islands C: the Hawaiian Islands D: the Galapagos Islands(\textcolor{gray}{\textit{Arc-Easy}})\\
        \hline
        \textbf{Output}: \textcolor[rgb]{0.9,0,0.1}{<plan\_start>Darwin's theory animal species change over time inspired research islands<plan\_end>}\textcolor{lightgray}{<paragraph>Speciation tendency of small, isolated genetic pools to produce unusual traits.  Examples include insular dwarfism and the radical changes among certain famous island chains, for example on Komodo.  The Galápagos Islands are particularly famous for their influence on Charles Darwin.  During his five weeks there he heard that Galápagos tortoises could be identified by island, and noticed that finches differed from one island to another, but it was only nine months later that he reflected that such facts could show that species were changeable.  When he returned to England, his speculation on evolution deepened after experts informed him that these were</paragraph>}\textcolor[rgb]{0,0.7,0.3}{<fparagraph>The Galápagos Islands are particularly famous for their influence on Charles Darwin.</fparagraph>}\textcolor[rgb]{1,0.65,0}{<answer\_start>D<answer\_end>}\\
        \hline
    \end{tabular}
    }
    \caption{Examples for short-form datasets}
    \label{tab:data short examples}
\end{table*}

\begin{table*}[t]
    \centering
    \resizebox{\textwidth}{!}{
    \begin{tabular}{p{17cm}}
        \hline
        \textbf{Input}: Answer the following question. The question may be ambiguous and have multiple correct answers, and in that case, you have to provide a long-form answer including all correct answers.\#\# Input: Where is superoxide dismutase found in the body?(\textcolor{gray}{\textit{ASQA}})\\
        \hline
        \textbf{Output}: \textcolor[rgb]{0.9,0,0.1}{<plan\_start>different types of superoxide dismutase found in the body<plan\_end>}\textcolor{lightgray}{<paragraph>Reactive oxygen species nearly all cells exposed to oxygen. In mammals and most chordates, three forms of superoxide dismutase are present. SOD1 is located primarily in the cytoplasm, SOD2 in the mitochondria and SOD3 is extracellular. The first is a dimer (consists of two units), while the others are tetramers (four subunits). SOD1 and SOD3 contain copper and zinc ions, while SOD2 has a manganese ion in its reactive centre. The genes are located on chromosomes 21, 6, and 4, respectively (21q22.1, 6q25.3 and 4p15.3-p15.1). The SOD-catalysed dismutation of superoxide may be written with the following half-reactions: where M = Cu ("n" =</paragraph>}\textcolor[rgb]{0,0.7,0.3}{<fparagraph>SOD1 is located primarily in the cytoplasm, SOD2 in the mitochondria and SOD3 is extracellular.The genes are located on chromosomes 21, 6, and 4, respectively (21q22.1, 6q25.3 and 4p15.3-p15.1).</fparagraph>}\textcolor[rgb]{1,0.65,0}{<answer\_start>Three forms of superoxide dismutase are present in humans, in all other mammals, and in most chordates, SOD1, SOD2, and SOD3, each located in a different part of the body.<answer\_end>}\textcolor[rgb]{0.9,0,0.1}{<plan\_start>different types of superoxide dismutase (SOD1, SOD2, SOD3) found in the body<plan\_end>}\textcolor{lightgray}{<paragraph>Reactive oxygen species nearly all cells exposed to oxygen. In mammals and most chordates, three forms of superoxide dismutase are present. SOD1 is located primarily in the cytoplasm, SOD2 in the mitochondria and SOD3 is extracellular. The first is a dimer (consists of two units), while the others are tetramers (four subunits). SOD1 and SOD3 contain copper and zinc ions, while SOD2 has a manganese ion in its reactive centre. The genes are located on chromosomes 21, 6, and 4, respectively (21q22.1, 6q25.3 and 4p15.3-p15.1). The SOD-catalysed dismutation of superoxide may be written with the following half-reactions: where M = Cu ("n" =</paragraph>}\textcolor[rgb]{0,0.7,0.3}{<fparagraph>SOD1 is located primarily in the cytoplasm, SOD2 in the mitochondria and SOD3 is extracellular.</fparagraph>}\textcolor[rgb]{1,0.65,0}{<answer\_start>Superoxide dismutase 1, or SOD1, is located in the cytoplasm, the material within a eukaryotic cell enclosed by the cell membrane except for the cell nucleus.<answer\_end>}\textcolor[rgb]{0.9,0,0.1}{<plan\_start>different types of superoxide dismutase (SOD1, SOD2, SOD3) found in the body<plan\_end>}\textcolor{lightgray}{<paragraph>Reactive oxygen species nearly all cells exposed to oxygen. In mammals and most chordates, three forms of superoxide dismutase are present. SOD1 is located primarily in the cytoplasm, SOD2 in the mitochondria and SOD3 is extracellular. The first is a dimer (consists of two units), while the others are tetramers (four subunits). SOD1 and SOD3 contain copper and zinc ions, while SOD2 has a manganese ion in its reactive centre. The genes are located on chromosomes 21, 6, and 4, respectively (21q22.1, 6q25.3 and 4p15.3-p15.1). The SOD-catalysed dismutation of superoxide may be written with the following half-reactions: where M = Cu ("n" =</paragraph>}\textcolor[rgb]{0,0.7,0.3}{<fparagraph>SOD1 is located primarily in the cytoplasm, SOD2 in the mitochondria and SOD3 is extracellular.SOD1 and SOD3 contain copper and zinc ions, while SOD2 has a manganese ion in its reactive centre.</fparagraph>}\textcolor[rgb]{1,0.65,0}{<answer\_start>SOD2 is found in the mitochondria, an organelle found in most eukaryotic organisms.<answer\_end>}\textcolor[rgb]{0.9,0,0.1}{<plan\_start>Superoxide dismutase SOD3 found in the body<plan\_end>}\textcolor{lightgray}{<paragraph>SOD3 SOD3 Extracellular superoxide dismutase  is an enzyme that in humans is encoded by the "SOD3" gene. This gene encodes a member of the superoxide dismutase (SOD) protein family. SODs are antioxidant enzymes that catalyze the dismutation of two superoxide radicals into hydrogen peroxide and oxygen. The product of this gene is thought to protect the brain, lungs, and other tissues from oxidative stress. The protein is secreted into the extracellular space and forms a glycosylated homotetramer that is anchored to the extracellular matrix (ECM) and cell surfaces through an interaction with heparan sulfate proteoglycan and collagen. A fraction of</paragraph>}\textcolor[rgb]{0,0.7,0.3}{<fparagraph>SOD3 SOD3 Extracellular superoxide dismutase  is an enzyme that in humans is encoded by the "SOD3" gene.The protein is secreted into the extracellular space and forms a glycosylated homotetramer that is anchored to the extracellular matrix (ECM) and cell surfaces through an interaction with heparan sulfate proteoglycan and collagen.</fparagraph>}\textcolor[rgb]{1,0.65,0}{<answer\_start>Finally, SOD3 is extracellular, meaning it occurs in the space outside the plasma membrane of a cell.<answer\_end>}\\
        \hline
    \end{tabular}
    }
    \caption{An Example for ASQA datasets}
    \label{tab:asqa data examples}
\end{table*}

\begin{table*}[t]
    \centering
    \resizebox{\textwidth}{!}{
    \begin{tabular}{p{17cm}}
        \hline
        \textbf{Input}: Which movie came out first Escape to Witch Mountain or Pete's Dragon ?(\textcolor{gray}{\textit{HotpotQA}})\\
        \hline
        \textbf{Output}: \textcolor[rgb]{0.9,0,0.1}{<plan\_start>Escape to Witch Mountain release<plan\_end>}\textcolor{lightgray}{<paragraph>Escape to Witch Mountain is a 1975 American fantasy-children's film, adapted from the 1968 science fiction novel of the same name written by Alexander H. Key.  The film was produced by Walt Disney Productions, released in March 1975 by Buena Vista Distribution Company and directed by John Hough. </paragraph>}\textcolor[rgb]{0,0.7,0.3}{<fparagraph>Escape to Witch Mountain is a 1975 American fantasy-children's film, adapted from the 1968 science fiction novel of the same name written by Alexander H. Key</fparagraph>}\textcolor[rgb]{1,0.65,0}{<answer\_start>Escape to Witch Mountain came out first,<answer\_end>}\textcolor[rgb]{0.9,0,0.1}{<plan\_start>Pete's Dragon release<plan\_end>}\textcolor{lightgray}{<paragraph>Pete's Dragon is a 2016 American fantasy comedy-drama adventure film directed by David Lowery, written by Lowery and Toby Halbrooks, and produced by James Whitaker.  The film is a live-action reimagining of Disney's 1977 live-action/animated musical film of the same name written by Malcolm Marmorstein.  The film stars Bryce Dallas Howard, Oakes Fegley, Wes Bentley, Karl Urban, Oona Laurence, and Robert Redford.  The film tells the story of an orphaned feral boy who befriends a dragon in the Pacific Northwest, and the ensuing repercussions of their discovery by the town's local residents. </paragraph>}\textcolor[rgb]{0,0.7,0.3}{<fparagraph>Pete's Dragon is a 2016 American fantasy comedy-drama adventure film directed by David Lowery, written by Lowery and Toby Halbrooks, and produced by James Whitaker. </fparagraph>}\textcolor[rgb]{1,0.65,0}{<answer\_start>before Pete's Dragon. <answer\_end>}[Combine]\textcolor[rgb]{1,0.65,0}{<answer\_start>Escape to Witch Mountain<answer\_end>}\\
        \hline
            \end{tabular}
            }
    \caption{An Example for HotpotQA datasets}
    \label{tab:hotpotqa data examples}
\end{table*}

% Additionally, we compared the impact of different fine-grained paragraph extractors on the RPG method, utilizing both GPT-3.5 and BGE-reranker for comparison. Results shown in Table~\ref{tab:extractor} indicates that the performance of both methods is very similar.
% \begin{table}[t]
%     \centering
%     \begin{tabular}{c|cc}
%     \hline
%     \textbf{Model}	& \textbf{ROUGE-L}	& \textbf{MAUVE} \\ \hline
% GPT-3.5 &	37.4 &	83.3 \\ \hline
% bge-reranker &	37.6 &	84.4 \\ \hline
%     \end{tabular}
%     \caption{Comparision of different extractor}
%     \label{tab:extractor}
% \end{table}